\pdfoutput=1
\documentclass[10pt]{article}
\usepackage[a4paper,textwidth=5.5in,textheight=9in,top=1in,headheight=0pt,headsep=0pt,footskip=0pt]{geometry}
\usepackage[T1]{fontenc}
\usepackage[utf8]{inputenc}
\usepackage{mathptmx}                 
\usepackage{courier}                  
\usepackage{textcomp}
\usepackage[safe]{tipa}                   
\usepackage{graphicx}
\DeclareUnicodeCharacter{025B}{\textepsilon}                 
\DeclareUnicodeCharacter{0254}{\textopeno}                   
\DeclareUnicodeCharacter{03B5}{\ensuremath{\varepsilon}}     
\DeclareUnicodeCharacter{1D10}{\reflectbox{\textsc{c}}}      
\usepackage{amsmath,amssymb}
\usepackage{graphicx,booktabs,array,caption,enumitem,xcolor}
\usepackage{xurl}
\usepackage{float}
\usepackage[hidelinks]{hyperref}
\setlist{itemsep=2pt,topsep=2pt}

\begin{document}
\thispagestyle{empty}

\begin{center}
\rule{\textwidth}{3pt}\\[10pt]
{\LARGE\bfseries MargEmbed-Akan: Bridging Twi Questions\\[4pt] and English Knowledge through\\[4pt] Cross-Lingual Score Distillation\par}
\vspace{8pt}\rule{\textwidth}{1pt}\\[22pt]
{\small\begin{tabular}{c@{\hspace{2.2em}}c@{\hspace{2.2em}}c}
\textbf{Prince N.\ Y.\ Annan} & \textbf{Ali A.\ A.\ Aziz} & \textbf{Stephen E.\ Moore}\\
University of Cape Coast & University of Cape Coast & University of Cape Coast\\
{\footnotesize\texttt{princeannan4332@gmail.com}} & {\footnotesize\texttt{azizalialhassan@gmail.com}} & {\footnotesize\texttt{stephen.moore@ucc.edu.gh}}
\end{tabular}}\\[12pt]
{\small Model: \url{https://huggingface.co/DexelLabs/MargEmbed-Akan}}
\end{center}
\vspace{6pt}

\begin{center}{\large\bfseries Abstract}\end{center}
\vspace{-4pt}
\begin{quote}
Most of the information a Twi (Akan) speaker looks for online is written in English, so a practical Twi retrieval system must match Twi questions to English passages. We study this setting in two parts. First, we evaluate eight retrieval systems, namely BM25, five open embedding models and two proprietary models (gemini-embedding-001 and OpenAI text-embedding-3-large), on two Twi--English tasks: AfriQA, in which 1,221 natively written Asante Twi questions must retrieve the English Wikipedia passage that answers them, and Bible translation retrieval, in which a Twi passage must retrieve its English translation among 6,195 candidates. Second, we introduce MargEmbed-Akan, a query encoder trained from AfriE5-Large-instruct by cross-lingual score distillation. The frozen base model scores each English training question against its answer passage and seven hard negatives; the student learns to reproduce those scores from an NLLB-200 Twi translation of the question, kept only if it survives a round-trip translation check, and from the English question itself, so that English behaviour is preserved. During training, passages are encoded by the frozen base model; at evaluation, MargEmbed-Akan encodes both questions and passages. Among existing systems, base AfriE5 and gemini-embedding-001 perform identically on AfriQA (66.6 nDCG@10), while EmbeddingGemma 2 and Qwen3-VL-Embedding-2B reach 56.6 and 51.3, and OpenAI text-embedding-3-large reaches 53.4. MargEmbed-Akan, a 560M-parameter open model, reaches 71.5 nDCG@10, significantly above gemini-embedding-001 (+4.9), its base model (+5.0) and OpenAI text-embedding-3-large (+18.1), with its largest gains on Twi typed without the letters ɛ and ɔ (+7.5 over gemini-embedding-001). Its English retrieval does not differ significantly from the base model, while both proprietary models remain 1.5--2.2 points better in English. On Bible translation retrieval, MargEmbed-Akan and gemini-embedding-001 are statistically tied on standard Twi (97.2 vs.\ 96.5 nDCG@10), and MargEmbed-Akan is clearly better on diacritic-free Twi (92.7 vs.\ 84.9). No AfriQA data was used for training or model selection.
\end{quote}

\section{Introduction}
Retrieval-augmented generation (RAG) grounds a language model's answer in passages retrieved for the user's question (Lewis et al., 2020); if the right passage is not retrieved, the answer cannot be grounded in it. Dense retrievers built on multilingual text embedding models now dominate this step, and benchmarks such as MTEB (Muennighoff et al., 2023) and MMTEB (Enevoldsen et al., 2025) guide which model practitioners choose. African languages remain underrepresented in these benchmarks: AfriMTEB (Uemura et al., 2026) notes that existing African-language tasks are often repurposed from translation benchmarks, and extends MMTEB to 59 languages.

Twi, a group of Akan dialects that is among the most widely spoken languages in Ghana, illustrates the practical problem. Much of the content a Twi speaker needs (news, health, education, government information) is written in English, so the realistic retrieval task is \emph{Twi question $\rightarrow$ English passage}. Twi orthography adds a second difficulty: the letters ɛ and ɔ are often typed as e and o on standard keyboards, or replaced by visually similar characters. In the AfriQA test split, 181 of 486 Twi questions use such look-alike characters, for example the Greek ε instead of ɛ.

AfriQA (Ogundepo et al., 2023) provides natively written Twi questions paired with English Wikipedia passages, which makes this task measurable. Using retrievers available in 2023, it found that translating the question into English before retrieval worked far better than retrieving directly with the Twi question. Whether current multilingual embedding models have closed that gap, and how an open model can be improved for Twi without Twi-native training data, are open questions. We ask three research questions:
\begin{enumerate}
\item \textbf{RQ1:} How well do current open and proprietary embedding models retrieve English passages for natively written Twi questions, and for Twi passages seeking their English translation?
\item \textbf{RQ2:} How sensitive are these models to how Twi is typed, that is, to look-alike characters and to missing ɛ and ɔ?
\item \textbf{RQ3:} Can an open embedding model be improved for Twi$\rightarrow$English retrieval using only machine-translated training questions, without degrading its English retrieval?
\end{enumerate}
Our contributions are:
\begin{itemize}
\item \textbf{A controlled evaluation} of BM25, multilingual-e5-large-instruct, AfriE5-Large-instruct, BGE-M3, EmbeddingGemma 2, Qwen3-VL-Embedding-2B, gemini-embedding-001 and OpenAI text-embedding-3-large on question$\rightarrow$passage retrieval (AfriQA) and translation retrieval (Bible), with three typing variants of each Twi query, paired bootstrap confidence intervals and multiple-comparison correction.
\item \textbf{MargEmbed-Akan}, a Twi query encoder trained by cross-lingual listwise score distillation with hard negatives, round-trip-filtered NLLB-200 translations, English anchoring and a frozen passage encoder, together with a fully documented data pipeline.
\item \textbf{Findings for practitioners:} the African-adapted AfriE5 already matches gemini-embedding-001 on native Twi questions and far exceeds OpenAI text-embedding-3-large; recent general-purpose models perform far worse on Twi than their English scores suggest; and MargEmbed-Akan significantly improves Twi$\rightarrow$English retrieval, especially for Twi typed without ɛ and ɔ, without a significant loss in English retrieval.
\end{itemize}

\section{Related Work}
\textbf{Embedding benchmarks.} BEIR (Thakur et al., 2021) established zero-shot evaluation of retrievers across heterogeneous datasets, and MTEB (Muennighoff et al., 2023) and MMTEB (Enevoldsen et al., 2025) broadened evaluation to many tasks and languages. AfriMTEB (Uemura et al., 2026) extends MMTEB for African languages to 59 languages, 14 tasks and 38 datasets, and introduces AfriE5, an adaptation of multilingual-e5-large-instruct (Wang et al., 2024) to African languages through cross-lingual contrastive distillation, trained on nine African languages that do not include Twi. We use AfriE5 as our base model and test how far its adaptation transfers to Twi.

\textbf{African-language question answering.} AfriQA (Ogundepo et al., 2023) is a cross-lingual open-retrieval question-answering dataset of 12,239 questions in ten African languages, written by native speakers and answered from English or French Wikipedia; its Twi questions use the Asante variety. With BM25 and an mBERT-based dense retriever, AfriQA found that translating the question first far outperformed direct cross-lingual retrieval: for Twi, Recall@10 was 21.4 with direct dense retrieval, 58.0 after Google Translate and 66.7 with human translation. AfriQA releases gold English passages, which we use for evaluation only.

\textbf{Cross-lingual distillation.} Reimers and Gurevych (2020) extend a monolingual sentence-embedding model to new languages by training a student so that both a source sentence and its translation map to the teacher's embedding of the source sentence. MargEmbed-Akan follows this teacher--student idea, including the step that keeps the student's source-language output close to the teacher, but distils retrieval scores rather than embedding vectors.

\textbf{Score distillation for retrieval.} Hofstätter et al.\ (2020) distil ranking knowledge by training a student to match the teacher's score margins between relevant and non-relevant passages (Margin-MSE), and RocketQAv2 (Ren et al., 2021) uses listwise distillation with soft relevance labels over candidate lists. Hard negatives, passages that are similar to the question but do not answer it, are central to training dense retrievers (Karpukhin et al., 2020). Our objective combines these ideas across languages: the teacher scores an English question against a candidate list containing hard negatives, and the student reproduces the score distribution from the Twi question.

\textbf{Weight averaging.} Interpolating the weights of a pre-trained and a fine-tuned model can retain the fine-tuned gains while recovering robustness lost in fine-tuning (Wortsman et al., 2022). We use this as a safeguard and select the interpolation weight on validation data.

\textbf{Twi resources.} GhanaNLP released an English--Twi parallel corpus of 25,421 sentence pairs in Akuapem Twi, with 697 crowd-sourced pairs for evaluation (Azunre et al., 2021). The eBible corpus (Akerman et al., 2023) provides openly licensed, verse-aligned Bible translations in 833 languages, including the Asante Twi and English translations we use for the translation-retrieval task.

\section{Evaluation Data}
We evaluate on two tasks that share the direction Twi $\rightarrow$ English but differ in what is being matched: a question with the passage that answers it, and a passage with its translation. Neither test set was used for training or model selection (Section~4.6).

\subsection{AfriQA: Twi question $\rightarrow$ English answer passage}
\textbf{Source.} We use the Twi portion of AfriQA (Ogundepo et al., 2023). Its questions were written by native speakers of Asante Twi and translated into English by the annotation team; annotators then identified an English Wikipedia paragraph containing the answer and marked the answer span. We use every Twi question with a released gold passage across the train, dev and test splits: \textbf{1,221 questions}, of which \textbf{486} belong to the test split. We report the test split and all splits, which gives more statistical power. Because MargEmbed-Akan was trained without any AfriQA data, its train and dev questions are unseen test questions for every system.

\textbf{Search corpus.} Searching all of English Wikipedia, as AfriQA does, was not feasible for many systems including rate-limited API models. We instead pooled every English gold passage that AfriQA releases for any language and split (Bemba, Hausa, Igbo, Kinyarwanda, Swahili, Twi, Yorùbá and Zulu), removed duplicates and empty passages, and prefixed each passage with its Wikipedia article title. This yields \textbf{5,357 passages} on African topics written in the same style as the gold passages, so each question competes against thousands of topically similar distractors. Absolute scores are therefore not comparable with those reported by AfriQA on the full Wikipedia corpus.

\textbf{Query versions.} Each Twi question is evaluated in three forms: \emph{as written}, exactly as typed by the annotator; \emph{standardized ɛ/ɔ}, with look-alike characters (Greek ε, ᴐ and similar) replaced by the standard letters; and \emph{without ɛ/ɔ}, with all variants mapped to e/o and tone marks removed, as when typing on a plain keyboard. The English translation of each question serves as a reference (Eng$\rightarrow$Eng), which is also the upper bound for a pipeline that translates the question perfectly before searching.

\textbf{Relevance.} \emph{Strict} scoring counts only the annotated gold passage and is our main measure. As a secondary measure we compute AfriQA's own criterion, under which any passage containing the English answer string also counts; because short answers occur in many passages, this measure is lenient and is reported only in the released tables.

\subsection{Bible translation retrieval: Twi passage $\rightarrow$ English translation}
\textbf{Source.} We use two verse-aligned translations from the eBible corpus (Akerman et al., 2023): Biblica's open Asante Twi Bible (2020, CC BY-SA 4.0) and the Berean Standard Bible (public domain). Both align line by line with a shared verse-reference list. We keep the 66-book Protestant canon, attach merged verses to the preceding verse, and group verses into non-overlapping windows of five within each chapter, merging a final window shorter than three verses into the previous one. This produces \textbf{6,195 parallel passages} of 3--7 verses.

\textbf{Task.} We draw \textbf{1,000 query passages} at random (seed 13). Each must retrieve its own translation among all 6,195 passages in the other language, so there is exactly one correct match. We evaluate Twi$\rightarrow$English, Twi without ɛ/ɔ$\rightarrow$English, and English$\rightarrow$Twi. Recall@1 is the most informative metric here, because a translation is either ranked first or confused with another passage.

\textbf{Why this task.} Translation retrieval does not depend on question style, answer annotation or a single gold paragraph among several valid ones; it directly measures whether a model places Twi and English text with the same meaning close together. It also lies outside the domain of MargEmbed-Akan's training data, which contains no Bible text. Because the Bible is widely available online, all models may have seen parts of it during pre-training.

\section{MargEmbed-Akan}
\subsection{Design}
MargEmbed-Akan is trained against a fixed passage space. During training, a frozen copy of AfriE5-Large-instruct acts as the \textbf{teacher} and encodes all passages (the targets); a trainable copy initialised from the same weights is the \textbf{student} and encodes the queries (the source). Keeping the passage side fixed gives the student a stable target and prevents its representations from drifting. After training, MargEmbed-Akan is used as a single model that encodes both questions and passages. Passage encoding was not trained directly, so we verified it empirically: on AfriQA, MargEmbed-Akan with its own passage embeddings scores 71.5 nDCG@10, against 71.3 when passages are encoded by AfriE5, a difference that is not significant. An existing AfriE5 passage index can therefore also be reused. Training requires no Twi-native data: we translate English training questions into Twi and train the student so that the Twi question ranks candidate passages the way the teacher ranks them for the original English question.

\subsection{Training questions: collection and processing}
\textbf{Sources.} We use four public English question$\rightarrow$passage collections in Sentence Transformers format (Table~\ref{tab:data}). Natural Questions contains real Google search queries answered by Wikipedia passages (Kwiatkowski et al., 2019); TriviaQA contains entity-centred trivia questions (Joshi et al., 2017), close in style to AfriQA; SQuAD questions were written by crowd workers while reading Wikipedia paragraphs (Rajpurkar et al., 2016); and GooAQ contains everyday questions collected through Google's autocomplete, with answers mined from Google answer boxes (Khashabi et al., 2021).

\textbf{Cleaning.} Each collection was processed in the same order: (1) Unicode NFC normalisation and whitespace clean-up; (2) removal of questions shorter than 10 characters and passages shorter than 30; (3) truncation of passages to their first 300 words, which fit within the base model's 512-token limit; (4) removal of duplicate questions; and (5) removal of duplicate passages, keeping one question per passage. The last step matters because candidate lists treat other passages as non-answers, so two questions sharing a passage would teach the model that a correct passage is wrong. It also explains most of the reduction for SQuAD, which has several questions per paragraph.

\textbf{Decontamination.} Before sampling, we removed any training question that matched an AfriQA question exactly after normalisation, or shared a six-word sequence containing at least two content words with any AfriQA question (in English or Twi) or with Bible-based questions from a preliminary benchmark. Generic phrases such as ``what is the capital of'' do not trigger the filter. This removed 808 questions (260 from Natural Questions, 470 from TriviaQA, 53 from SQuAD and 25 from GooAQ). The training sources contain no Bible text.

\textbf{Keyword-overlap sampling.} For each pair we computed keyword overlap: the share of the question's content words that also appear in the passage. The median overlap of each cleaned pool was 0.60--0.71, so training mostly on such pairs would reward word matching, which an English passage index cannot offer a Twi question. We therefore over-sampled low-overlap pairs, targeting 60\% low ($\le$0.3), 30\% medium and 10\% high ($>$0.6) overlap, with per-source targets of 15,000, 8,000, 5,000 and 12,000 pairs. High-overlap pairs were kept deliberately, because real questions often name the entity they ask about. Natural Questions and SQuAD contained fewer low-overlap pairs than targeted, and all of them were used. The selected set is 55\% low, 34\% medium and 11\% high overlap.

\begin{table}[!htbp]
\centering\small
\caption{Training data at each processing stage (question--passage pairs).}\label{tab:data}
\resizebox{\textwidth}{!}{%
\begin{tabular}{lcrrrr}
\toprule
Source & Columns & Pool & Selected & After filters & After round trip\\
\midrule
Natural Questions & \texttt{query}/\texttt{answer} & 74,909 & 12,673 & 12,580 & 12,511\\
TriviaQA & \texttt{query}/\texttt{answer} & 63,658 & 8,000 & 7,964 & 7,952\\
SQuAD & \texttt{question}/\texttt{answer} & 18,834 & 3,126 & 3,121 & 3,121\\
GooAQ (first 300k rows) & \texttt{question}/\texttt{answer} & 230,558 & 12,000 & 11,964 & 11,911\\
\midrule
\textbf{Total} & & \textbf{387,959} & \textbf{35,799} & \textbf{35,629} & \textbf{35,495}\\
\bottomrule
\end{tabular}}
\end{table}

\subsection{Translation and round-trip filtering}
Questions were translated from English (\texttt{eng\_Latn}) into Twi (\texttt{twi\_Latn}) with NLLB-200 3.3B (NLLB Team, 2022), using beam search with two beams and at most 96 new tokens. Outputs that were empty, identical to the source, of implausible length (character ratio outside 0.5--3) or repeating one word four or more times were discarded. Because a mistranslated question would teach the student to give a wrong Twi question the scores of the original, we applied a \textbf{round-trip check}: each Twi translation was translated back into English, the teacher embedded the original and the back-translated question, and the pair was kept only if their cosine similarity was at least 0.80. This kept 35,495 of 35,629 questions (99.6\%); median similarities per source were 0.94--0.97. The rejected outputs were clearly broken; the lenient threshold mainly removed severe failures rather than subtle ones (Section~8).

\subsection{Teacher scores and hard negatives}
For each English question, the teacher computed cosine similarities between the question (with its retrieval instruction) and all $\sim$35,500 training passages. From the teacher's top 30 we removed the question's own passage and any passage scoring at least 97\% of the answer passage's score, since such passages are likely also correct, and sampled seven \textbf{hard negatives} from the rest. Each question therefore has eight candidates, and the teacher's eight similarity scores are the training targets.

\subsection{Objective}
For a training question with candidates $c_1,\dots,c_8$, the frozen teacher $T$ scores the English question $q_{en}$, and the student $S$ scores the version being trained on, $q$, which is either the Twi translation $q_{tw}$ or the English question itself. Both distributions are compared after a softmax with temperature $\tau=0.05$:
\begin{equation}
t_i=\cos\!\big(T(q_{en}),T(c_i)\big),\quad s_i=\cos\!\big(S(q),T(c_i)\big),\quad
\mathcal{L}_{\text{distill}}=\mathrm{KL}\big(\mathrm{softmax}(t/\tau)\,\big\|\,\mathrm{softmax}(s/\tau)\big).
\end{equation}
The loss is computed for $q=q_{tw}$ (Twi$\rightarrow$English learning) and for $q=q_{en}$ (English$\rightarrow$English, with equal weight). The English term anchors the student to the teacher's English behaviour and prevents English drift. Because the student matches a score distribution over candidates rather than the teacher's full embedding vector, it is free to place Twi questions anywhere that preserves the teacher's ranking. A light alignment term uses 5,000 sentence pairs from the GhanaNLP English--Twi dataset on Hugging Face: the student's embeddings of the Twi and English sentence are pulled towards the teacher's embedding of the English sentence (cosine distance, weight 0.5). To reflect how Twi is typed, we added diacritic-free copies of 25\% of the Twi training questions and look-alike-character copies (ɛ$\rightarrow$ε, ɔ$\rightarrow$ᴐ) of 15\%, with the same targets, giving 47,945 distillation rows.

\subsection{Training and model selection}
The student was trained for two epochs (412 steps) with batches of 256 questions, learning rate $1\times10^{-5}$ with linear decay and 10\% warm-up, 8-bit AdamW (Dettmers et al., 2022), bfloat16 mixed precision and gradient checkpointing, with questions truncated to 128 tokens during training and 512 at inference. Training took 33.7 minutes on one NVIDIA L4 GPU.

\textbf{Validation without AfriQA.} We held out 1,000 translated questions with their candidate lists and measured, every 50 steps, how highly the student ranked the answer among the eight candidates (MRR) for the Twi and English versions. A natively written Twi validation set was not available, so selection used translated questions. Twi MRR rose from 0.773 before training to 0.871, while English MRR stayed at about 0.987. The training loss fell from about 0.070 at the first logged step to about 0.035 at step 412. The final checkpoint scored best on the held-out validation set.

\textbf{Weight averaging.} Following Wortsman et al.\ (2022), we compared interpolations of the trained and base weights with 30\%, 50\%, 70\% and 100\% trained weights, requiring that English MRR drop by at most one point. All four blends passed and Twi MRR increased monotonically (0.820, 0.838, 0.857, 0.871), so the unblended trained model was selected as MargEmbed-Akan.

\section{Experimental Setup}
\textbf{Systems.} Table~\ref{tab:systems} lists the systems. All were run in a single evaluation (proprietary models through their APIs) with identical queries and corpora. MargEmbed-Akan encodes both questions and passages. For AfriQA, instruction-following models received ``Given a question, retrieve Wikipedia passages that answer the question''; for Bible translation, ``Given a Bible passage, retrieve the passage that is its translation''. Gemini used \texttt{RETRIEVAL\_QUERY}/\texttt{RETRIEVAL\_DOCUMENT} for AfriQA and \texttt{SEMANTIC\_SIMILARITY} for translation. Retrieval is exact cosine search over the full corpus. BM25 (Robertson and Zaragoza, 2009) is applied only where query and passages share a language.

\begin{table}[H]
\centering\small
\caption{Systems evaluated.}\label{tab:systems}
\resizebox{\textwidth}{!}{%
\begin{tabular}{lll}
\toprule
System & Type & Query prompt\\
\midrule
BM25 & Keyword baseline (English queries only) & --\\
multilingual-e5-large-instruct (Wang et al., 2024) & Open, 560M & E5 instruction\\
AfriE5-Large-instruct (Uemura et al., 2026) & Open, 560M & E5 instruction\\
\textbf{MargEmbed-Akan (ours)} & Open, 560M & E5 instruction\\
BGE-M3 (Chen et al., 2024) & Open, 568M & none\\
EmbeddingGemma 2 (Google DeepMind, 2026) & Open & named task prompts\\
Qwen3-VL-Embedding-2B (Li et al., 2026) & Open, 2B & instruction\\
gemini-embedding-001 (Lee et al., 2025) & Proprietary API & task types\\
text-embedding-3-large (OpenAI, 2024) & Proprietary API & none\\
\bottomrule
\end{tabular}}
\end{table}

\textbf{Metrics.} The main metric is nDCG@10; with one relevant passage it equals $1/\log_2(r+1)$ when the gold passage is at rank $r\le10$, and 0 otherwise. We also report Recall@1 and Recall@$k$.

\textbf{Statistics.} We compute 95\% confidence intervals with a paired bootstrap over questions (10,000 resamples; Koehn, 2004). Comparisons of MargEmbed-Akan with the proprietary models and with its base model were planned in advance; their p-values are corrected with the Holm (1979) procedure within this family.

\section{Results}
\subsection{Question $\rightarrow$ passage retrieval (AfriQA)}
\begin{figure}[!htbp]
\centering
\includegraphics[width=\textwidth]{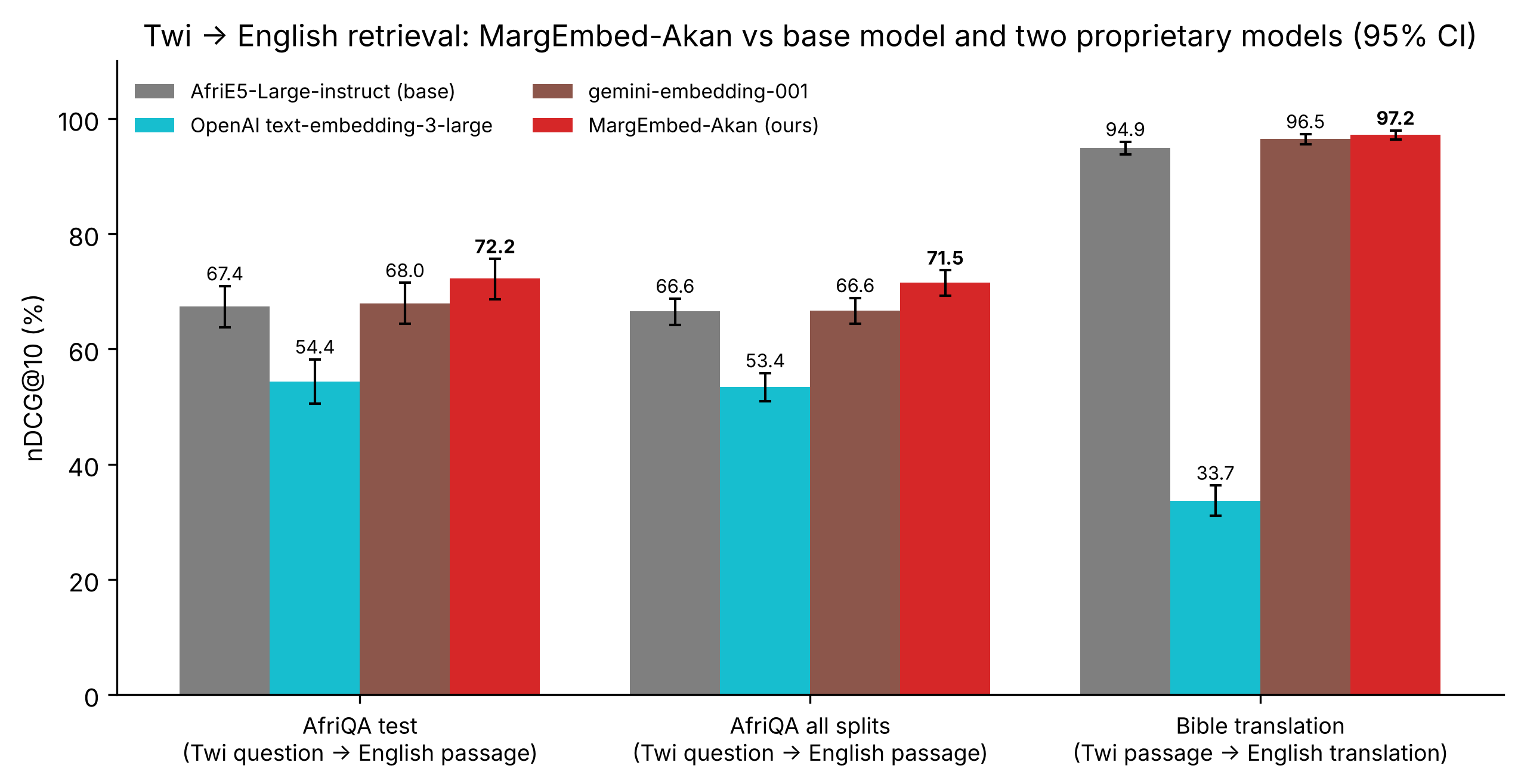}
\caption{Twi $\rightarrow$ English nDCG@10 on the three test sets for base AfriE5, the two proprietary models and MargEmbed-Akan (95\% confidence intervals).}\label{fig:headline}
\end{figure}

\begin{figure}[!htbp]
\centering
\includegraphics[width=0.85\textwidth]{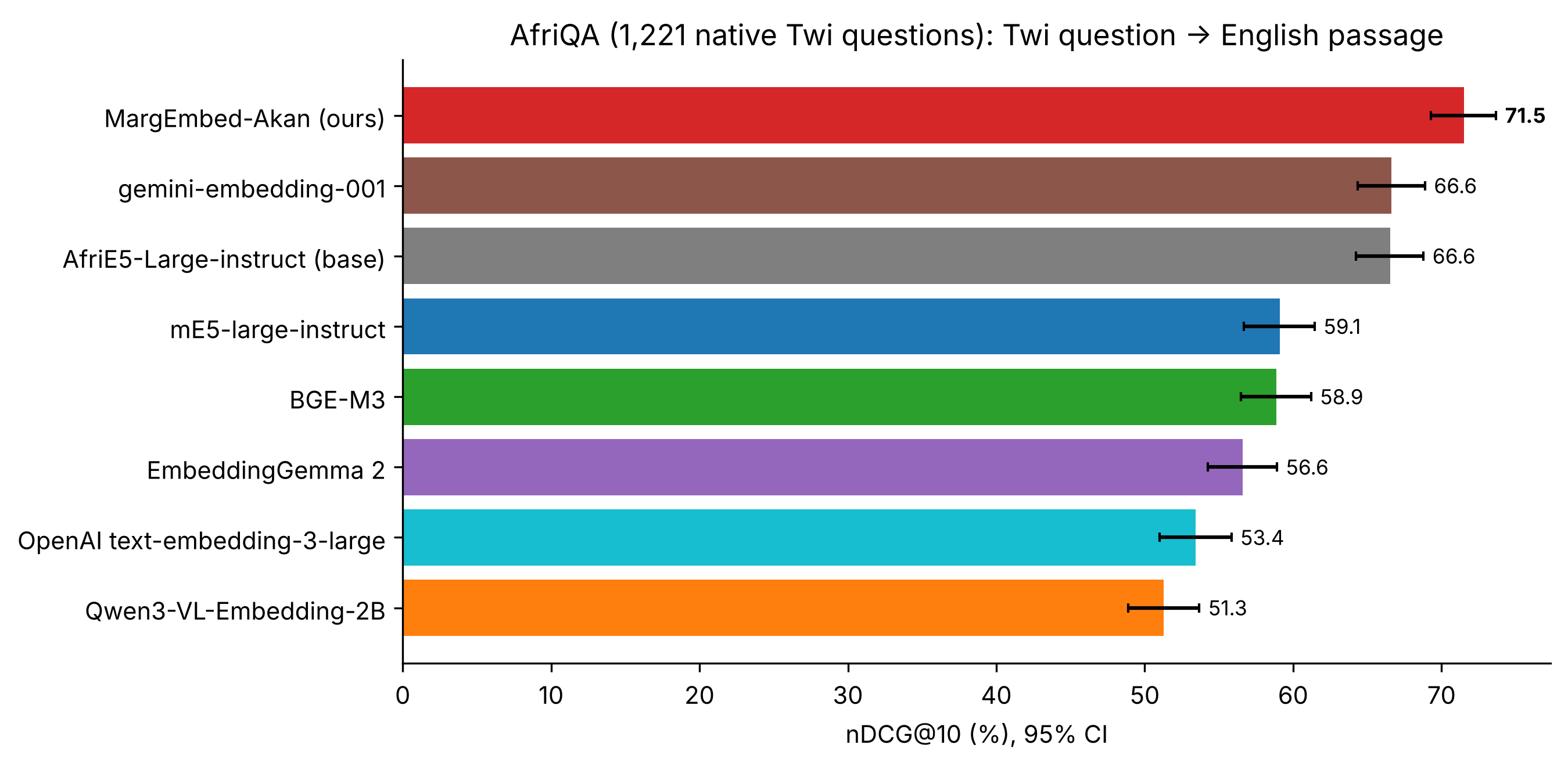}
\caption{All embedding systems on AfriQA (all splits), Twi question $\rightarrow$ English passage.}\label{fig:ranking}
\end{figure}

MargEmbed-Akan is the strongest system in every Twi setting (Table~\ref{tab:afriqa}, Figures~\ref{fig:headline} and~\ref{fig:ranking}). Among existing models, AfriE5 and gemini-embedding-001 are tied (66.6), well ahead of the other open models and of OpenAI text-embedding-3-large (53.4). The ranking changes completely in English: all embedding models score 82--87, and the two proprietary models are best. Asking in Twi rather than English costs every system between 13.2 (MargEmbed-Akan) and 32.8 points (text-embedding-3-large). The improvement is largest at the top of the ranking: MargEmbed-Akan places the gold passage first for 60.9\% of questions, against 54.1\% for gemini-embedding-001 and 53.8\% for AfriE5. Results on the 486-question test split follow the same pattern (Table~\ref{tab:test}).

\begin{table}[!htbp]
\centering\small
\caption{AfriQA, all 1,221 Twi questions: nDCG@10 (\%) with 95\% confidence intervals. Best Twi result per column in bold.}\label{tab:afriqa}
\resizebox{\textwidth}{!}{%
\begin{tabular}{lcccc}
\toprule
System & Twi as written & Twi, standard ɛ/ɔ & Twi without ɛ/ɔ & English (reference)\\
\midrule
\textbf{MargEmbed-Akan (ours)} & \textbf{71.5} [69.3, 73.7] & \textbf{71.5} [69.3, 73.7] & \textbf{69.5} [67.3, 71.7] & 84.7 [83.1, 86.3]\\
gemini-embedding-001 & 66.6 [64.4, 68.9] & 68.7 [66.4, 71.0] & 62.0 [59.7, 64.3] & 86.9 [85.4, 88.3]\\
AfriE5-Large-instruct (base) & 66.6 [64.2, 68.8] & 67.6 [65.3, 69.8] & 61.5 [59.0, 63.8] & 85.2 [83.5, 86.7]\\
mE5-large-instruct & 59.1 [56.7, 61.4] & 60.6 [58.2, 62.9] & 53.8 [51.4, 56.2] & 84.1 [82.5, 85.8]\\
BGE-M3 & 58.9 [56.5, 61.2] & 58.8 [56.4, 61.1] & 55.2 [52.8, 57.6] & 84.2 [82.6, 85.9]\\
EmbeddingGemma 2 & 56.6 [54.3, 58.9] & 56.7 [54.3, 59.0] & 55.8 [53.4, 58.1] & 85.1 [83.5, 86.6]\\
OpenAI text-embedding-3-large & 53.4 [51.0, 55.8] & 53.9 [51.4, 56.3] & 53.0 [50.6, 55.4] & 86.2 [84.6, 87.8]\\
Qwen3-VL-Embedding-2B & 51.3 [48.9, 53.7] & 51.4 [49.1, 53.8] & 49.4 [47.0, 51.8] & 82.4 [80.6, 84.1]\\
BM25 & -- & -- & -- & 69.2 [67.0, 71.4]\\
\bottomrule
\end{tabular}}
\end{table}

\begin{table}[!htbp]
\centering\small
\caption{AfriQA test split (486 questions): nDCG@10 (\%).}\label{tab:test}
\begin{tabular}{lccc}
\toprule
System & Twi as written & Twi without ɛ/ɔ & English (reference)\\
\midrule
\textbf{MargEmbed-Akan (ours)} & \textbf{72.2} & \textbf{70.4} & 86.9\\
gemini-embedding-001 & 68.0 & 63.0 & 89.6\\
AfriE5-Large-instruct (base) & 67.4 & 63.2 & 87.7\\
OpenAI text-embedding-3-large & 54.4 & 53.9 & 89.4\\
\bottomrule
\end{tabular}
\end{table}

\subsection{Significance of the main comparisons}
MargEmbed-Akan significantly outperforms both proprietary models on Twi questions, as written and without ɛ/ɔ, in every AfriQA set, and its base model by 3.9--8.1 points (Table~\ref{tab:sig}). Its English retrieval does not differ significantly from the base model; gemini-embedding-001 and text-embedding-3-large are 1.5--2.6 points better in English. On Bible translation retrieval, MargEmbed-Akan and gemini-embedding-001 are statistically tied for Twi$\rightarrow$English, while MargEmbed-Akan leads significantly for English$\rightarrow$Twi and for Twi without ɛ/ɔ.

\begin{table}[!htbp]
\centering\small
\caption{Difference in nDCG@10 (MargEmbed-Akan minus comparison system) with 95\% bootstrap confidence intervals. * significant after Holm correction over the planned comparisons.}\label{tab:sig}
\resizebox{\textwidth}{!}{%
\begin{tabular}{llccc}
\toprule
Test set & Setting & vs gemini-embedding-001 & vs text-embedding-3-large & vs AfriE5 (base)\\
\midrule
AfriQA test & Twi as written & +4.3 [+1.7, +6.9]* & +17.9 [+14.7, +21.0]* & +4.8 [+3.1, +6.6]*\\
AfriQA test & Twi without ɛ/ɔ & +7.4 [+4.5, +10.3]* & +16.5 [+13.5, +19.7]* & +7.2 [+5.1, +9.4]*\\
AfriQA test & English & $-$2.6 [$-$4.3, $-$1.0]* & $-$2.4 [$-$4.1, $-$0.7]* & $-$0.7 [$-$1.7, +0.1]\\
AfriQA all splits & Twi as written & +4.9 [+3.1, +6.6]* & +18.1 [+16.1, +20.1]* & +5.0 [+3.9, +6.1]*\\
AfriQA all splits & Twi without ɛ/ɔ & +7.5 [+5.6, +9.4]* & +16.5 [+14.5, +18.5]* & +8.1 [+6.7, +9.5]*\\
AfriQA all splits & English & $-$2.2 [$-$3.3, $-$1.1]* & $-$1.5 [$-$2.7, $-$0.5]* & $-$0.5 [$-$1.0, +0.0]\\
Bible translation & Twi $\rightarrow$ English & +0.8 [$-$0.3, +1.8] & +63.5 [+60.9, +66.0]* & +2.3 [+1.4, +3.2]*\\
Bible translation & Twi without ɛ/ɔ $\rightarrow$ English & +7.8 [+5.8, +9.8]* & +59.7 [+57.1, +62.3]* & +19.2 [+17.2, +21.2]*\\
Bible translation & English $\rightarrow$ Twi & +2.5 [+1.3, +3.7]* & +65.6 [+63.0, +68.2]* & +1.5 [+0.8, +2.3]*\\
\bottomrule
\end{tabular}}
\end{table}

\textbf{Passage-overlap check.} Training questions were filtered against AfriQA questions, but training passages were not filtered against AfriQA passages. We therefore compared every passage in the cleaned training pools (387,959 passages, of which only 35,495 were used) with the AfriQA corpus, counting a match when two passages share a sequence of 13 or more words. 206 of the 1,221 Twi questions (84 of 486 in the test split) have a gold passage with such a match. Removing these questions leaves MargEmbed-Akan's scores essentially unchanged (71.6 nDCG@10 on the remaining 1,015 questions, 72.2 on the remaining 402 test questions). On the 1,015 questions it still leads gemini-embedding-001 by 4.4 points [+2.5, +6.4], text-embedding-3-large by 17.4 and AfriE5 by 4.7; on the 402 test questions its lead over gemini-embedding-001 is 3.2 points [+0.5, +6.1]. Overlap with training passages therefore does not explain the results.

\subsection{Robustness to how Twi is typed}
\begin{figure}[!htbp]
\centering
\includegraphics[width=\textwidth]{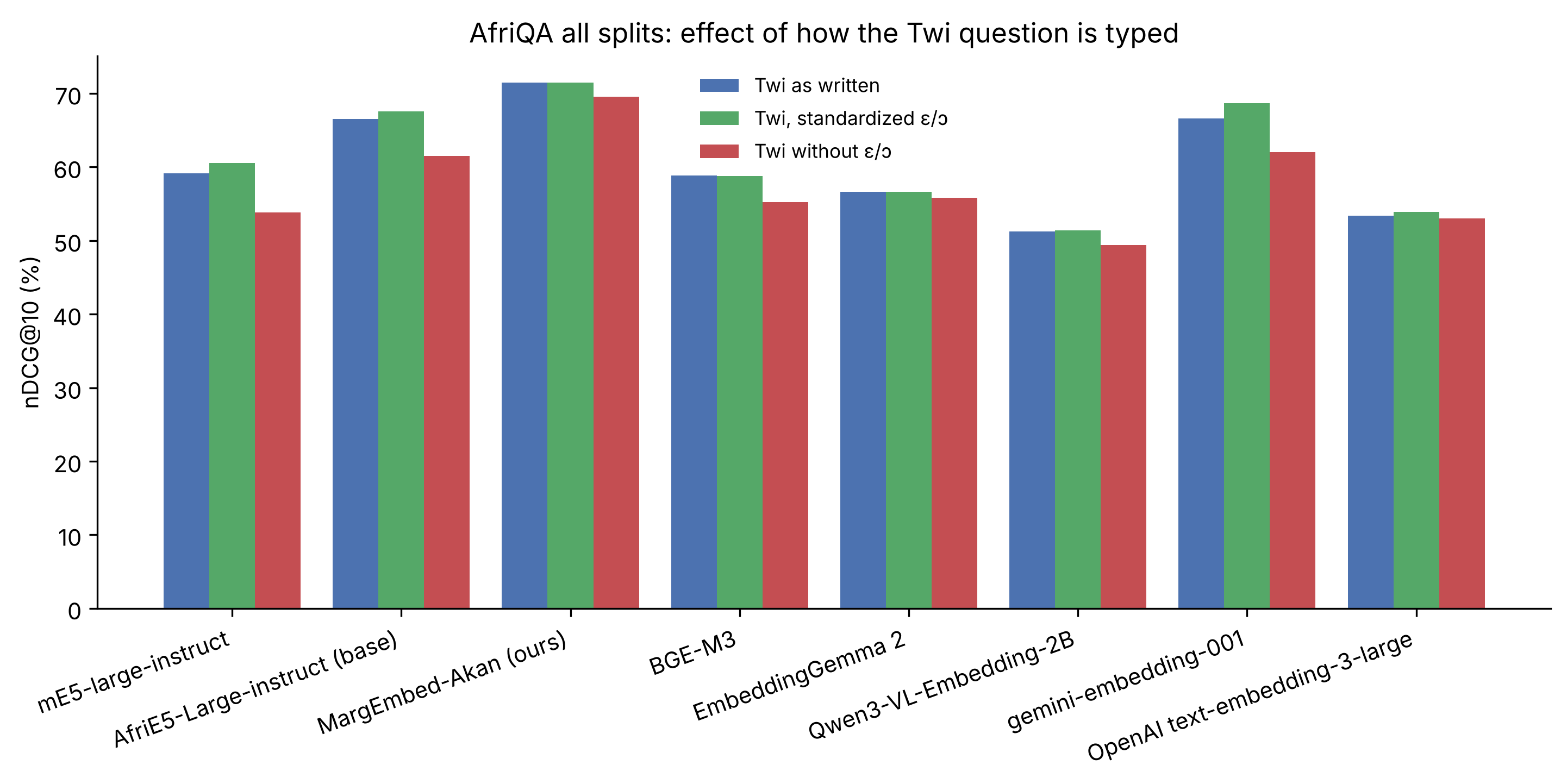}
\caption{AfriQA (all splits) nDCG@10 for Twi as written, with standardized ɛ/ɔ, and without ɛ/ɔ.}\label{fig:robust}
\end{figure}

Replacing look-alike characters with standard ɛ/ɔ helps models that were not trained on such variants, for example gemini-embedding-001 (+2.1 points), but makes no difference for MargEmbed-Akan, which saw look-alike variants during training (Figure~\ref{fig:robust}). Removing ɛ and ɔ costs gemini-embedding-001 4.6 points and AfriE5 5.1 points on AfriQA, but MargEmbed-Akan only 2.0. The effect is larger for translation retrieval: without ɛ/ɔ, the Bible Twi$\rightarrow$English score falls by 11.6 points for gemini-embedding-001 and 21.4 for AfriE5, but only 4.5 for MargEmbed-Akan. text-embedding-3-large changes little across variants because its Twi scores are already low.

\subsection{Translation retrieval (Bible)}
\begin{figure}[!htbp]
\centering
\includegraphics[width=\textwidth]{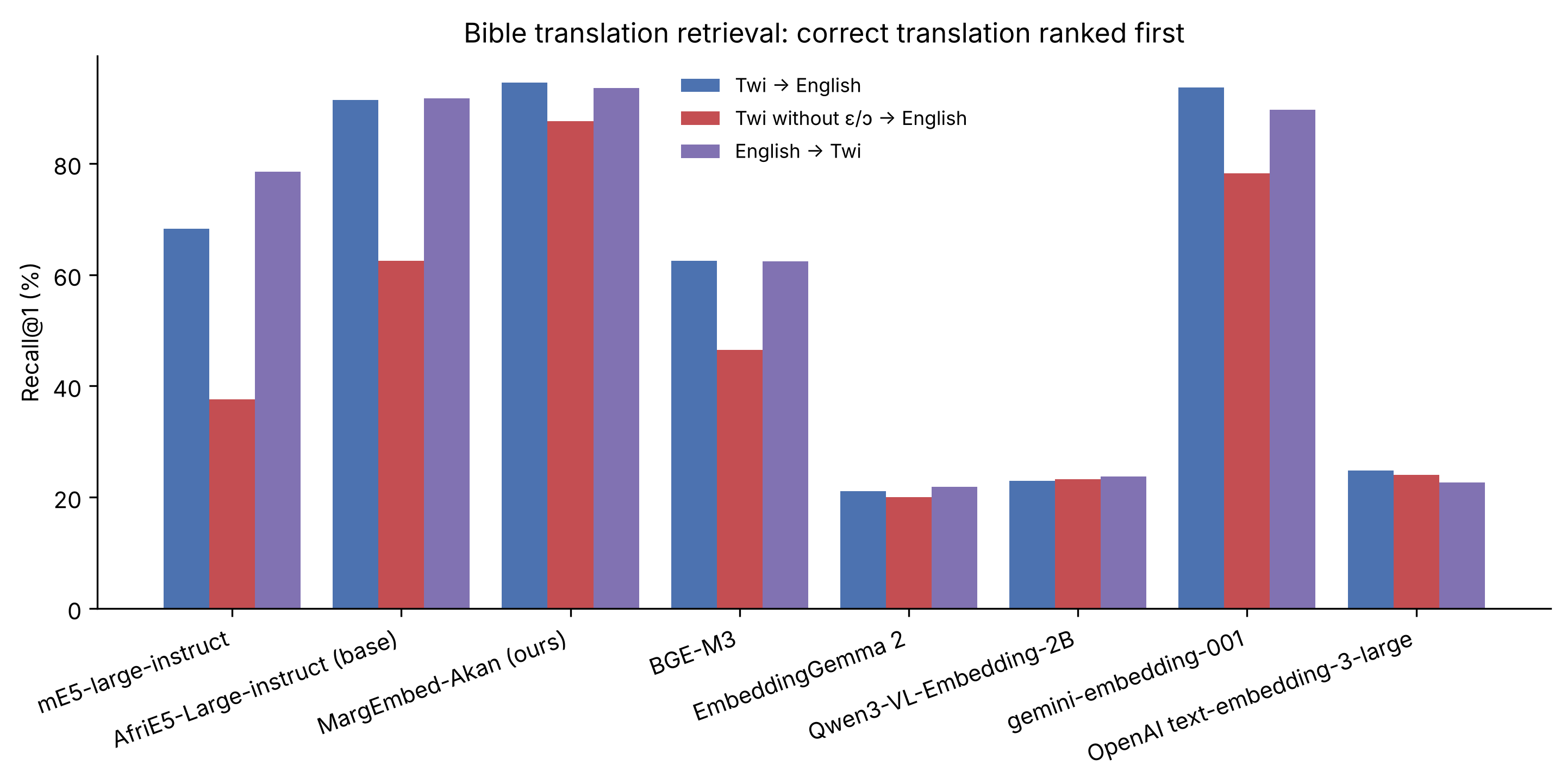}
\caption{Bible translation retrieval: share of passages whose translation is ranked first.}\label{fig:bible}
\end{figure}

The three strongest systems find the exact translation for over 89\% of passages, and MargEmbed-Akan improves on its base model although it was never trained on Bible text (Table~\ref{tab:bible}, Figure~\ref{fig:bible}). By contrast, text-embedding-3-large, Qwen3-VL-Embedding-2B and EmbeddingGemma 2 find the correct translation for fewer than 25\% of passages, showing that their Twi and English representations are poorly aligned despite strong English performance.

\begin{table}[!htbp]
\centering\small
\caption{Bible translation retrieval: Recall@1 (\%), 1,000 query passages, 6,195 candidates.}\label{tab:bible}
\begin{tabular}{lccc}
\toprule
System & Twi $\rightarrow$ English & Twi without ɛ/ɔ $\rightarrow$ English & English $\rightarrow$ Twi\\
\midrule
\textbf{MargEmbed-Akan (ours)} & \textbf{94.6} & \textbf{87.6} & \textbf{93.6}\\
gemini-embedding-001 & 93.7 & 78.2 & 89.7\\
AfriE5-Large-instruct (base) & 91.4 & 62.5 & 91.7\\
mE5-large-instruct & 68.3 & 37.6 & 78.5\\
BGE-M3 & 62.5 & 46.5 & 62.4\\
OpenAI text-embedding-3-large & 24.8 & 24.0 & 22.6\\
Qwen3-VL-Embedding-2B & 22.9 & 23.2 & 23.7\\
EmbeddingGemma 2 & 21.1 & 20.0 & 21.9\\
\bottomrule
\end{tabular}
\end{table}

\subsection{Latency}
We measured retrieval speed on a single NVIDIA T4 GPU in half precision, using 300 AfriQA Twi test questions and the 5,357-passage corpus (Table~\ref{tab:latency}). For a single question, MargEmbed-Akan returns the top-10 English passages in 17.8 ms (median; 23.1 ms at the 95th percentile), almost all of it spent encoding the question; exact search takes 0.16 ms. Batching raises throughput from 55 to 480 questions per second. The passage index in this test was built with AfriE5, which has the same architecture as MargEmbed-Akan, so indexing cost and latency are identical for either model, and unlike the proprietary models it runs locally without API calls or usage costs.

\begin{table}[H]
\centering\small
\caption{MargEmbed-Akan retrieval latency on one NVIDIA T4 GPU.}\label{tab:latency}
\begin{tabular}{lr}
\toprule
Measure & Result\\
\midrule
Indexing, 5,357 passages (one-off) & 37.3 s\\
Query encoding, p50 / p95 & 17.7 / 22.9 ms\\
Search (top-10), p50 / p95 & 0.16 / 0.18 ms\\
End-to-end per question, p50 / p95 & 17.8 / 23.1 ms\\
Throughput, batch size 1 / 8 / 32 / 128 & 55 / 351 / 480 / 428 questions/s\\
\bottomrule
\end{tabular}
\end{table}

\section{Discussion}
\textbf{Language coverage matters more than model size or recency.} The two proprietary models behave very differently on Twi: gemini-embedding-001 matches the African-adapted AfriE5, while text-embedding-3-large falls 13 points behind it on AfriQA and fails at translation retrieval. The most recent open models, EmbeddingGemma 2 and Qwen3-VL-Embedding-2B, show the same weakness. Strong English or aggregate multilingual scores are therefore a poor guide for Twi.

\textbf{Why MargEmbed-Akan works.} Three design choices explain its behaviour. First, training against a frozen passage space with an English distillation term keeps the model's English behaviour close to the base model, which is why MargEmbed-Akan can also encode passages without loss. Second, distilling the teacher's scores over hard-negative candidates teaches fine distinctions between similar passages, which shows up as the largest gains at rank 1. Third, augmenting training questions with diacritic-free and look-alike spellings makes the model robust to how people actually type Twi. Gains on Bible translation, a domain absent from training, suggest that the model improved its Twi--English alignment rather than fitting the training domain.

\textbf{Lessons from earlier attempts.} Before MargEmbed-Akan we tried two simpler approaches, each evaluated in separate runs: contrastive fine-tuning of AfriE5 on synthetic Twi data, and distillation of full embedding vectors (MSE) combined with in-batch retrieval training on NLLB-1.3B translations. Both improved robustness to missing ɛ/ɔ but not standard Twi retrieval, and both reduced English retrieval by 3--16 nDCG@10 points. Distilling ranking scores with a frozen passage encoder removed this trade-off.

\textbf{Practical recommendations.} For a Twi RAG system over English content, use MargEmbed-Akan for both questions and passages; an existing AfriE5 passage index can also be kept with equivalent results. For English-only search, gemini-embedding-001 or text-embedding-3-large remain slightly better. Even the best system loses 13.2 points when the question is asked in Twi rather than English, so substantial headroom remains.

\section{Limitations}
\textbf{Machine-translated training data.} All Twi training questions are NLLB-200 translations; the round-trip check removed only clearly broken outputs, so subtler errors remain. NLLB-200 is released under a non-commercial licence, so MargEmbed-Akan should be treated as a research model.
\textbf{No native-Twi validation.} Checkpoints and the weight-averaging ratio were selected on translated questions.
\textbf{Single run and no ablations.} MargEmbed-Akan was trained once with one seed, and we did not isolate the contribution of each component. Passage encoding was not trained directly; its quality is established by evaluation only.
\textbf{Evaluation scope.} The AfriQA corpus contains 5,357 passages rather than all of English Wikipedia, so absolute scores are not comparable with those reported by AfriQA; each question has one gold passage; Bible text may appear in pre-training data; and the proprietary models were accessed in October 2026 and may change.
\textbf{Passage overlap.} A 13-word overlap check found 206 AfriQA questions whose gold passage resembles a passage in the training pools; results excluding them are essentially unchanged (Section~6.2).
\textbf{Design iterations.} Earlier attempts were evaluated on the AfriQA test split; reporting all splits and the Bible task partly offsets this.
\textbf{Dialect.} The evaluation data is Asante Twi, while the GhanaNLP alignment corpus is largely Akuapem Twi; other Akan varieties are untested.

\section{Ethical Considerations}
All evaluation data is publicly released under open licences (AfriQA: CC BY 4.0; Asante Twi Bible: CC BY-SA 4.0; Berean Standard Bible: public domain), and no personal data was collected. Retrieval quality for Twi directly affects whether Twi speakers receive grounded answers from RAG systems; we release the model so that Ghanaian developers can build on it, and we caution that machine-translated training data may carry translation biases.

\section{Conclusion}
We evaluated eight retrieval systems on Twi$\rightarrow$English question answering and translation retrieval, and introduced MargEmbed-Akan, a query encoder trained by cross-lingual score distillation from AfriE5-Large-instruct using only machine-translated questions. Language coverage, not model size, determined performance on Twi. MargEmbed-Akan, a 560M-parameter open model, outperformed both proprietary models on natively written Twi questions (+4.9 nDCG@10 over gemini-embedding-001 and +18.1 over text-embedding-3-large on AfriQA), was most robust to Twi typed without ɛ and ɔ, and kept English retrieval close to its base model. Future work should add a native-Twi validation set, ablate each training component, and extend the approach to other Akan varieties and Ghanaian languages.

\textbf{Data and model availability.} The model is available at \url{https://huggingface.co/DexelLabs/MargEmbed-Akan}; it encodes both questions and passages, and an index built with \texttt{McGill-NLP/AfriE5-Large-instruct} can also be used. The evaluation data is public: AfriQA (CC BY 4.0) and the eBible corpus (Asante Twi Bible under CC BY-SA 4.0; Berean Standard Bible in the public domain).

\section*{References}
{\small\setlength{\parskip}{3pt}
\hangindent=1.5em Akerman, V., Baines, D., Daspit, D., Hermjakob, U., Jang, T., Leong, C., Martin, M., Mathew, J., Robie, J., and Schwarting, M. (2023). The eBible corpus: Data and model benchmarks for Bible translation for low-resource languages. \emph{arXiv:2304.09919}.

\hangindent=1.5em Azunre, P., Osei, S., Addo, S., et al. (2021). English-Twi parallel corpus for machine translation. \emph{AfricaNLP Workshop at EACL 2021}. arXiv:2103.15625.

\hangindent=1.5em Chen, J., Xiao, S., Zhang, P., Luo, K., Lian, D., and Liu, Z. (2024). M3-Embedding: Multi-linguality, multi-functionality, multi-granularity text embeddings through self-knowledge distillation. \emph{Findings of ACL 2024}.

\hangindent=1.5em Dettmers, T., Lewis, M., Shleifer, S., and Zettlemoyer, L. (2022). 8-bit optimizers via block-wise quantization. \emph{ICLR 2022}.

\hangindent=1.5em Enevoldsen, K., et al. (2025). MMTEB: Massive multilingual text embedding benchmark. \emph{ICLR 2025}.

\hangindent=1.5em Google DeepMind (2026). EmbeddingGemma 2 model card. \url{https://huggingface.co/google/embeddinggemma-2}.

\hangindent=1.5em Hofstätter, S., Althammer, S., Schröder, M., Sertkan, M., and Hanbury, A. (2020). Improving efficient neural ranking models with cross-architecture knowledge distillation. \emph{arXiv:2010.02666}.

\hangindent=1.5em Holm, S. (1979). A simple sequentially rejective multiple test procedure. \emph{Scandinavian Journal of Statistics}, 6(2):65--70.

\hangindent=1.5em Joshi, M., Choi, E., Weld, D.\ S., and Zettlemoyer, L. (2017). TriviaQA: A large scale distantly supervised challenge dataset for reading comprehension. \emph{Proceedings of ACL 2017}.

\hangindent=1.5em Karpukhin, V., Oğuz, B., Min, S., Lewis, P., Wu, L., Edunov, S., Chen, D., and Yih, W. (2020). Dense passage retrieval for open-domain question answering. \emph{Proceedings of EMNLP 2020}.

\hangindent=1.5em Khashabi, D., Ng, A., Khot, T., Sabharwal, A., Hajishirzi, H., and Callison-Burch, C. (2021). GooAQ: Open question answering with diverse answer types. \emph{Findings of EMNLP 2021}, 421--433.

\hangindent=1.5em Koehn, P. (2004). Statistical significance tests for machine translation evaluation. \emph{Proceedings of EMNLP 2004}.

\hangindent=1.5em Kwiatkowski, T., Palomaki, J., Redfield, O., et al. (2019). Natural Questions: A benchmark for question answering research. \emph{Transactions of the ACL}, 7:452--466.

\hangindent=1.5em Lee, J., Chen, F., Dua, S., et al. (2025). Gemini Embedding: Generalizable embeddings from Gemini. \emph{arXiv:2503.07891}.

\hangindent=1.5em Lewis, P., Perez, E., Piktus, A., et al. (2020). Retrieval-augmented generation for knowledge-intensive NLP tasks. \emph{Advances in Neural Information Processing Systems 33}.

\hangindent=1.5em Li, M., et al. (2026). Qwen3-VL-Embedding and Qwen3-VL-Reranker. \emph{arXiv:2601.04720}.

\hangindent=1.5em Muennighoff, N., Tazi, N., Magne, L., and Reimers, N. (2023). MTEB: Massive text embedding benchmark. \emph{Proceedings of EACL 2023}.

\hangindent=1.5em NLLB Team (2022). No language left behind: Scaling human-centered machine translation. \emph{arXiv:2207.04672}.

\hangindent=1.5em Ogundepo, O., Gwadabe, T.\ R., Rivera, C.\ E., et al. (2023). AfriQA: Cross-lingual open-retrieval question answering for African languages. \emph{Findings of EMNLP 2023}.

\hangindent=1.5em OpenAI (2024). New embedding models and API updates. \url{https://openai.com/index/new-embedding-models-and-api-updates/}.

\hangindent=1.5em Rajpurkar, P., Zhang, J., Lopyrev, K., and Liang, P. (2016). SQuAD: 100,000+ questions for machine comprehension of text. \emph{Proceedings of EMNLP 2016}.

\hangindent=1.5em Reimers, N.\ and Gurevych, I. (2020). Making monolingual sentence embeddings multilingual using knowledge distillation. \emph{Proceedings of EMNLP 2020}, 4512--4525.

\hangindent=1.5em Ren, R., Qu, Y., Liu, J., Zhao, W.\ X., She, Q., Wu, H., Wang, H., and Wen, J.-R. (2021). RocketQAv2: A joint training method for dense passage retrieval and passage re-ranking. \emph{Proceedings of EMNLP 2021}.

\hangindent=1.5em Robertson, S.\ and Zaragoza, H. (2009). The probabilistic relevance framework: BM25 and beyond. \emph{Foundations and Trends in Information Retrieval}, 3(4):333--389.

\hangindent=1.5em Thakur, N., Reimers, N., Rücklé, A., Srivastava, A., and Gurevych, I. (2021). BEIR: A heterogeneous benchmark for zero-shot evaluation of information retrieval models. \emph{NeurIPS 2021 Datasets and Benchmarks Track}.

\hangindent=1.5em Uemura, K., Zhang, M., and Adelani, D.\ I. (2026). AfriMTEB and AfriE5: Benchmarking and adapting text embedding models for African languages. \emph{Proceedings of EACL 2026}.

\hangindent=1.5em Wang, L., Yang, N., Huang, X., Yang, L., Majumder, R., and Wei, F. (2024). Multilingual E5 text embeddings: A technical report. \emph{arXiv:2402.05672}.

\hangindent=1.5em Wortsman, M., Ilharco, G., Kim, J.\ W., et al. (2022). Robust fine-tuning of zero-shot models. \emph{Proceedings of CVPR 2022}, 7959--7971.
\par}

\appendix
\section{Training Configuration}
\begin{table}[H]
\centering\small
\resizebox{\textwidth}{!}{%
\begin{tabular}{ll}
\toprule
Setting & Value\\
\midrule
Base model (teacher and student initialisation) & McGill-NLP/AfriE5-Large-instruct\\
Translation model & facebook/nllb-200-3.3B, beam size 2, max 96 new tokens\\
Round-trip threshold & cosine $\ge$ 0.80 (teacher embeddings)\\
Candidates per question & 1 answer + 7 hard negatives from teacher top 30\\
Distillation loss & listwise KL, $\tau$ = 0.05, Twi and English queries weighted equally\\
Alignment loss & cosine distance, weight 0.5, 5,000 GhanaNLP sentence pairs\\
Augmentation & 25\% diacritic-free, 15\% look-alike-character copies\\
Training rows & 47,945 distillation + 5,000 alignment\\
Optimiser and schedule & 8-bit AdamW, lr $1\times10^{-5}$, linear decay, 10\% warm-up\\
Batch size, epochs, steps & 256, 2, 412\\
Sequence length & 128 tokens (training), 512 (inference)\\
Precision and hardware & bfloat16, gradient checkpointing, 1 $\times$ NVIDIA L4, 33.7 min\\
Selection & held-out translated questions (Twi MRR); weight averaging, chosen $\alpha$ = 1.0\\
\bottomrule
\end{tabular}}
\end{table}

\section{Query Instructions}
For AfriQA: ``Given a question, retrieve Wikipedia passages that answer the question''. For Bible translation: ``Given a Bible passage, retrieve the passage that is its translation''. E5-style models use the prefix ``Instruct: \ldots\ Query: ''. One AfriQA question has an empty Twi text; all systems received it unchanged, except the Gemini API, which rejects empty input and received a single hyphen.

\end{document}